# Grading the Graders: Verification Autonomy Levels (L0–L5) for LLM Reasoning

**Yajie Yin**

Email: 1549080929@qq.com · ORCID: 0009-0001-6168-2530

## Abstract

Large language models (LLMs) are increasingly paired with "verifiers"—step checkers, self-consistency filters, tool-based fact checkers, and formal proof assistants—that claim to detect the model's errors. Yet the verification literature uses the word *level* to mean at least five different things: verification granularity, concept abstraction, risk tier, system-stack layer, and the epistemic source of the ground truth. We propose **Verification Autonomy Levels (VAL)**, a meta-standard that classifies any verification scheme along a single axis: *where does the verification spec come from, and what does the verdict guarantee?* VAL ranges from L0 (LLM self-declaration; no deterministic anchor) through L2 (objective ground truth; correctness only) to L3/L4 (decidable systems with single-property or domain-level completeness), with L5 shown to be impossible in the unrestricted case. Central to VAL is the **completeness blind spot**: substitution- and sampling-based verifiers can confirm that proposed candidates hold, but cannot prove that no candidate was missed. We further identify a dichotomy the literature has not stated: completeness is reachable only for *formally specifiable* properties, whereas empirical open-world verification (fact-checking, diagnosis) caps at anchored correctness (L2). We document this gap empirically across four domains—symbolic mathematics, behavior monitoring, medical diagnosis, and code generation (the fourth a *reverse validation*: framework predictions stated before evidence)—and in the strongest formal-verification baseline in our survey, whose own authors note the verifier "focuses on the correctness of each step." We further show that the levels of granularity, concept hierarchy, risk, and system stack are orthogonal to VAL, resolving a systematic conflation across 17 surveyed papers. We release a runnable classifier (`val_standard.py`) and the full literature assessment on Zenodo (DOI: 10.5281/zenodo.23120985).

---

## 1. Introduction

Large language models (LLMs) produce fluent, confident, and frequently wrong reasoning. The dominant mitigation is *verification*: attach a second mechanism that checks the model's claims. The 2023–2025 literature has produced a large and rapidly growing body of verification proposals—trained step verifiers [3], self-checking schemas [5], tool-augmented fact checkers [17], graph-structured verification [2], and formal proof assistants [4]—each claiming to catch the errors the model cannot self-report. This proliferation raises a question the literature has not explicitly asked: **what can a given verification scheme actually guarantee, and where does that guarantee come from?**

Classical verification theory explains how to verify a program against a *given* specification—the specification is assumed exogenous and trusted. LLM-based systems introduce a prior problem that this assumption excludes: **whether the specification, the evaluator, and the verification authority are themselves appropriately anchored.** We argue that this prior problem has a natural graded structure (VAL, Sec. 3), and that the LLM-verification literature has been answering it implicitly, without naming it or giving it an operational decision procedure.

We further identify a dichotomy the literature has not stated: completeness is achievable only for *formally specifiable* properties—those with a decidable fragment (mathematical, syntactic, rule-scoped)—whereas empirical open-world verification (fact-checking, diagnosis, legal analysis) has no such fragment and, for the open-world question, caps at anchored correctness (L2); at best it reaches rule-scoped completeness over a formalized sub-fragment, which remains silent about the open world. This is not a limitation of our framework but a property of the domain, and it is a central finding of this paper.

Answering this question is harder than it appears, because the field uses the word *level* to mean at least five different things:

1. **Granularity**—how finely the output is decomposed for checking: claim → sentence → document [7]; atomic step → paragraph [2].
2. **Concept abstraction**—the mathematical sophistication at which verification operates: foundational elements → high-level concepts [8].
3. **Risk/disposition**—what the verdict triggers: safe/unsafe/conditional tiers [10]; output-determinism tiers [12].
4. **System stack**—which component is audited: model → workflow → system [13]; data → base → execute → service [14].
5. **Epistemic anchoring**—the source of the ground truth a verdict rests on, and the strength of the guarantee it delivers.

Axes 1–4 answer *what* to check, *how finely*, *at which layer*, and *what to do with the result*. None answers the question that determines whether a verifier can ever be trusted: **on what ground truth does the verdict rest, and does it guarantee correctness, completeness, or neither?**

We argue that the fifth axis is the one that matters for trust, that it has a natural six-level structure, and that the literature's conflation of the five axes has obscured it. Three observations motivate the framework.

*First, the anchor is what fails.* In our four-domain study (Sec. 5), every verification failure we catalogued was traceable to the anchor, not to the judge: a well-calibrated verifier (42/42 unit cases) was fed wrong subproblems by the LLM it was checking (decomposition contamination), and never saw the final answer it should have checked (combination tampering), or was asked to verify a condition declared by the very model under test (trust recursion). A perfectly calibrated instrument pointed at the wrong object is not a bug; it is a specification problem.

*Second, correctness is not completeness*. The most common verifiers—substitution, sampling, statistical thresholds—can confirm that *proposed* candidates hold, but cannot prove that *no* candidate was missed. We call this the **completeness blind spot** and document it empirically in symbolic mathematics (a missed root that substitution cannot see), in behavior monitoring (a covert-execution attack that leaves no trace in the output layer), and in medical diagnosis (confident conclusions from insufficient evidence). The blind spot is not a tuning failure; it is a property of the verification paradigm (Sec. 4).

*Third, the strongest existing verification concedes the same point*. The most rigorous scheme in our survey—Lean 4 step-level formal verification [4]—states that its formal verifier "focuses on the correctness of each step": the kernel checks proofs of LLM-generated statements, but the statements themselves, and the case coverage they encode, remain LLM declarations. The blind spot does not disappear at the top of the ladder; it is pushed to the statement layer.

**Contributions.** We make four:

1. **Verification Autonomy Levels (VAL)**, a six-level epistemic taxonomy (L0–L5) classifying any verification scheme by its anchor source and guarantee, together with a deterministic decision procedure (Sec. 3) and a runnable classifier (`val_standard.py`).
2. **A disambiguation of five confounded "level" axes**, showing that granularity, concept abstraction, risk, and system-stack are orthogonal to the VAL axis, and locating 17 representative papers in the resulting space (Sec. 2).
3. **A formal and empirical treatment of the completeness blind spot**, including a statement of why universal completeness is impossible in the unrestricted case (Sec. 4.5) and why completeness is always relative to an operational design domain (Sec. 4).
4. **Four cross-domain case studies**—symbolic mathematics, behavior monitoring, medical diagnosis, and code generation, the last a *reverse validation* with predictions stated before evidence (Sec. 5)—that exercise the framework end-to-end and report honest negative results.

**Why this matters.** For a deployer, VAL is a pre-purchase checklist: ask *where the spec comes from* before trusting any verification claim, and know that an L2 verdict is a correctness probe, not a completeness guarantee. For a researcher, VAL separates the five questions hidden inside "hierarchical verification," preventing category errors such as claiming that finer granularity implies stronger grounding. For a reviewer, VAL supplies a vocabulary for interrogating any claim of the form "our system verifies X": *at what level, within what ODD, with what abstention behavior?*

**Positioning.** We present VAL as a conceptual contribution. The four case studies of Sec. 5 are exercises of the framework, not claims of empirical superiority over baselines—indeed, two report clean negative results (the verification architecture does not improve accuracy in mathematics or code generation). The empirical content is intentionally modest; the claim we defend is structural: that one axis—the source of the verification spec—is measurable, ordinal, and currently untheorized in the LLM-verification literature.

---

## 2. Related Work: Five Axes, One Word

We reviewed 17 representative papers spanning what the literature calls "layered" or "hierarchical" verification, and classified each along five axes. **Evidence boundary: the assessment is abstract-level for 11 of these 17 papers; the 6 anchor-axis papers were verified against full text (Limitation 1).** The full assessment—per-paper evidence, confidence levels, and a versioned re-verification trail—is released as supplementary material (`docs/07`); here we report the axis structure and the anchor-axis classifications used throughout.

### 2.1 Granularity axis: *how fine*

Graph of Verification [2] adapts verification granularity from atomic steps (formal tasks) to whole paragraphs (informal narratives) via a "node block" architecture, trading precision against robustness. Factcheck-GPT [7] annotates factuality at three granularities—claim, sentence, document—with a GPT-4-based annotation scheme and a gold-labeled benchmark. Dr. V [9] decomposes video-hallucination diagnosis into perceptual, temporal, and cognitive levels, grounding the first two in 10k spatial-temporal gold annotations. These are ladders of *decomposition fineness*. They say nothing about the anchor: the same granularity ladder can be implemented with LLM judgment (L0) or with objective ground truth (L2).

### 2.2 Concept axis: *how abstract*

Hierarchical Attention [8] regularizes LLM attention toward a five-level hierarchy of mathematical concepts to improve proof generation in formal theorem proving (miniF2F, ProofNet). Its proofs are checked by a formal kernel—an L4 anchor—but the paper's contribution is a generation-time regularizer, and its "levels" are concept-abstraction levels, orthogonal to the anchor.

### 2.3 Risk/disposition axis: *what to do*

A safety-response framework [10] classifies inputs into four disposition tiers (Safe, Unsafe, Conditionally Safe, Focused Attention) via a supervised fine-tuned classifier, reporting 99.3% recall. LLM Output Drift [12] tiers models by output determinism for risk-adapted deployment in finance, combining consistency measurement with invariant checking and SEC-citation validation. $M^3$-SafetyBench [11] evaluates models across content- and functional-safety dimensions with a 170k-item benchmark. These are ladders of *consequence*—what the verdict should trigger. A "four-tier safety classifier" and a "three-level fact-checking pipeline" are frequently cited together, yet one is a disposition ladder and the other a granularity ladder; neither is an epistemic ladder.

### 2.4 System-stack axis: *which layer*

Standard Benchmarks Fail [13] proposes stress-testing financial LLM agents at model, workflow, and system layers, arguing that standard accuracy benchmarks "provide an illusion of reliability." Prompting Frameworks Survey [14] organizes prompting tooling into data, base, execute, and service layers. These are ladders of *audit scope*. [13] is notable as the one paper in our survey that makes completeness of evaluation coverage an explicit thesis.

### 2.5 Anchor axis: *on what ground truth* (this paper)

Classified by the procedure of Sec. 3.3:

- **L0** — SelfCheck [5]: four-stage regenerate-and-compare, explicitly "without resorting to external resources"; the checker itself scores 66.7% verification accuracy and the authors concede "the checks are themselves imperfect." $LM^2$ [6]: a verifier language model, fine-tuned on GPT-4 annotations and coordinated with the decomposer and solver via policy learning.
- **L0/L1 + tools** — VerifiAgent [1]: meta-verification of completeness and consistency is LLM-based with some deterministic rules (L0/L1); tool-based adaptive verification delegates factual and computational checks to a Python interpreter, a search engine, and symbolic computation—deterministic execution serving as *silver ground truth*, a reference weaker than gold (L1), as the authors confirm (personal communication).
- **L0/L1** — Factcheck-GPT [7]: runtime verdicts delegated to a GPT-4-based annotation scheme; the gold labels in its benchmark anchor the benchmark, not the runtime verdict (corrected from L1/L2 at abstract-level re-verification, Sec. 2.6).
- **L1/L2** — FACTOOL [17]: tools (Google Search, Google Scholar, code interpreters) gather evidence, but the final factuality verdict is LLM reasoning over that evidence. LLM Output Drift [12]: consistency measurement plus invariant checking with SEC-citation validation.
- **L2** — DiVERSE [3]: a DeBERTa-v3 step verifier trained on step labels derived by matching against ground-truth answers. $M^3$-SafetyBench [11]: a gold-labeled evaluation benchmark. Dr. V [9]: hallucination diagnosis grounded in gold spatial-temporal annotations.
- **L3/L4** — Hierarchical Attention [8]: proofs checked by a formal kernel. Safe [4]: Lean 4 step verification—the kernel is L4, but the theorem statements are LLM-generated, i.e., L0 at the statement layer.

| Axis | Question it answers | Representative papers |
|---|---|---|
| **Anchor (VAL)** | What ground truth does the verdict rest on, with what guarantee? | SelfCheck L0; $LM^2$ L0; VerifiAgent L0/L1+tools; DiVERSE L2; FACTOOL L1/L2; Safe L4+L0 |
| Granularity | How finely is the output decomposed for checking? | GoV, Factcheck-GPT, Dr. V |
| Concept | At what abstraction level does verification operate? | Hierarchical Attention |
| Risk/Disposition | What response does the verdict trigger? | SafetyResponse, OutputDrift, $M^3$-SafetyBench |
| System stack | Which component of the system is audited? | BenchmarksFail, PromptingSurvey |

**Observation.** Across all 17 papers, none formalizes the anchor as a ladder, none treats the completeness of a verification scheme itself as an object of study, and the single most explicit acknowledgment of the gap comes from the strongest formal baseline [4]. The conflation is not harmless: it lets "layered verification" papers inherit each other's credibility across axes that do not entail one another. The rest of this paper develops the anchor axis.

### 2.6 Adjacent fields: formal methods, risk frameworks, epistemology

Readers from program analysis and formal methods will recognize the correctness/completeness distinction as decades old: Hoare logic distinguishes partial from total correctness; abstract interpretation formalizes the soundness and completeness of abstractions; type theory defines completeness via logical relations. We agree, and state precisely what VAL adds.

In traditional program verification the division of labor is fixed: a human writes the specification, and the verifier proves the program against it. The specification is ground truth *by construction*—it does not depend on the program under analysis. In LLM verification this division collapses: the "program" (the model's reasoning) and the "specification" (what should be checked) are both produced by LLMs. The anchor-source axis—whether the spec is LLM-declared (L0), derived from the problem text (L1), grounded in objective truth (L2), or encoded in a decidable system (L3/L4)—is the new object of study. Hoare logic tells us how to verify a program against a given specification; VAL asks *who gave the specification and whether it can be trusted*. The former is soundness theory; the latter is an accountability taxonomy for a class of artifacts in which the spec is no longer exogenous.

The distinction from risk-management and maturity frameworks (NIST AI RMF, CMMI) is different: those are disposition ladders—they classify what to do about risk (Sec. 2.3's risk axis), not the epistemic status of a verdict. A system can be process-mature and still run L0 verification; maturity and anchor strength are orthogonal. Epistemic philosophy (reliabilism, justification theory) and LLM interpretability frameworks [16] ask whether a belief is *justified* or how it was formed; VAL does not adjudicate truth—it classifies the structural source of a claim's checkability. Adjacent surveys of trustworthy evaluation [15] map this territory without isolating the anchor axis. We do not claim to resolve epistemology; we claim that one axis—spec source—is ordinal, operationally decidable (Sec. 3.3), and currently absent from the LLM-verification literature.

Closest in the security-assurance tradition are the Common Criteria Evaluation Assurance Levels (EAL1–EAL7, grading the assurance of the evaluation *process*), DO-178C Design Assurance Levels (DAL A–E, grading by failure *consequence*), and the trusted computing base (TCB, delimiting which components must be *trusted*). These grade process maturity, consequence, or trust scope; none grades the *source of the verification specification*. VAL's axis is the fourth question: given a verification claim, who declared what is to be verified, and does that declaration carry an independent anchor? This is the question the security-assurance ladders do not answer, and it is the one that matters most when the verifier and the verified share a generator (Sec. 3.2).

**Auditability of this assessment.** The literature assessment in this paper is itself versioned and auditable (supplementary material, docs/07): initial judgments (V1.0), abstract-level verification (V1.2), and full-text verification of the six anchor-axis papers (V1.3) are recorded as a change log. One judgment was corrected at abstract-level re-verification (V1.2): a paper initially classified L1/L2 was found to delegate its verdict to a GPT-4-based judge, L0/L1; full-text verification of the six anchor-axis papers (V1.3) subsequently upheld every remaining judgment. We report this not as an embarrassment but as the framework applied to itself: assessments are claims, claims require anchors, and the correction trail is the anchor.

---

## 3. The VAL Framework

### 3.1 Levels

| Level | Anchor source | Guarantee | Completeness | Driving analogy |
|---|---|---|---|---|
| **L0** | LLM self-declaration | none | none | full manual (the driver "confidently errs") |
| **L1** | deterministic rules derived from the problem text/code | deterministic matching | no | lane keeping (single-function assist) |
| **L2** | objective ground truth / gold labels / *independent* oracle (not a sibling generator) | correctness | no | partial automation (human must take over) |
| **L3** | definitional/provable (property encoded in a decidable system) | single-property **completeness** | yes (within ODD) | conditional automation (system owns liability in ODD) |
| **L4** | domain-level proof systems | domain-wide completeness | yes (within domain) | high automation (no takeover in ODD) |
| **L5** | universal completeness | any property | **impossible** (unrestricted) | full automation—does not exist |

Each level is defined by two properties: the **anchor source** (who or what supplies the ground truth the verdict rests on) and the **guarantee** (what a PASS commits to). Three consequences follow.

First, *the anchor, not the judge, determines the level.* A perfectly calibrated verifier at L2 is still L2: substitution checking that a candidate satisfies an equation proves the candidate holds; it says nothing about candidates not proposed. Improving the *implementation* of an L2 check (denser sampling, a better threshold) moves the system horizontally within L2; only changing the anchor source moves it vertically.

Second, *the guarantee degrades downward but not upward.* A judge built for L3 (e.g., a symbolic solver) can be *used* at L2 (substitution only)—we call this a *usage-degraded* anchor, and the classifier flags it as upgradeable. A judge built for L2 cannot be promoted to L3 by more data: no amount of sampling closes a completeness gap (Sec. 4). This asymmetry is why "more data" is not a level-raising operation.

Third, *every level has a characteristic abstention behavior, and it is diagnostic.* L0 systems deny error —they "confidently err." L1/L2 systems are silent about what they did not check. L3/L4 systems return a decidable abstention when a property falls outside their ODD (Safe's "failed formalization" state is exactly this [4]). L5 does not exist. In our experience, a verifier's abstention behavior is a faster diagnostic of its level than any benchmark score.

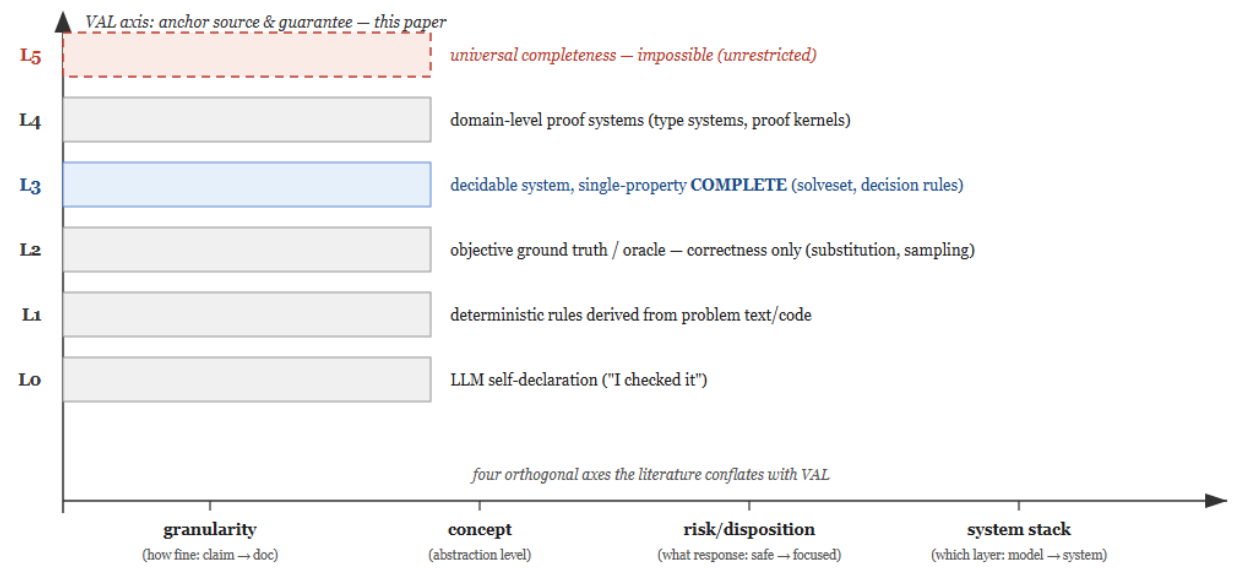

**Figure 1.** The VAL ladder (vertical) and four orthogonal axes (horizontal) that the literature conflates with it. Granularity, concept, risk, and system-stack answer *what / how fine / what response / which layer*; VAL answers *on what ground truth, with what guarantee.*

## 3.2 The three decisive questions

Any verification scheme is classified by answering three questions in order:

- **Q1 (spec source).** Who declares the verification condition—the thing the verdict is about? The answer is one of: (a) the LLM under test itself ("I checked it," "this is verified"); (b) a deterministic rule derived from the problem or code text (regex, parser, schema); (c) an objective source independent of both the problem and the generative mechanism under test (gold answer, measured value, execution result); (d) a property encoded in a decidable system (a symbolic solver, a type system, a decision rule with validated thresholds).

**Anchor independence.** "Objective" in L2 does not mean merely "external to the model under test." An anchor qualifies only if it is *independent of the generative mechanism being verified*—its error mechanism must not be correlated with the verified system's. An answer produced by another LLM is external but not independent: it shares the error-generating mechanism, and is L0 masked as L2. Execution results, symbolic simplification, and physical measurements qualify because their "error mechanism" is the artifact itself, not a correlated generator. This is the operational form of trust-recursion termination (Sec. 6.1): the anchor chain stops at definitional or conventional anchors, never at a sibling generator. Objective anchors are not all equal; they split by what they certify. A *gold-label anchor* (gold ground truth)—a known-correct reference—certifies the answer itself and is L2. A *silver anchor* (silver ground truth)—deterministic re-computation that serves as a reference—certifies the computation, not the answer, and sits at the L1/L2 boundary: objective about whether the computation reproduces a result, silent about whether that result is correct. The term *silver ground truth* is due to the authors of VerifiAgent, whose tool component is exactly this—it re-derives with a Python interpreter or symbolic engine and compares—which is why the author places it at L1 rather than L2 (Sec. 2.5).

**Anchor semantics.** Gold anchors further split by *what they certify*, and the distinction changes what a verdict means. An **intent anchor** certifies a recorded decision event—that a user or platform authorized an action toward a target (a confirmation record, an approval). A **truth anchor** certifies a fact—that an answer matches objective ground truth (gold labels, measured values, a verified equation). An **effect anchor** certifies an execution outcome—that an action actually occurred or succeeded (server-side state change, test execution). The same L2 level can host all three, but they are not interchangeable: an intent anchor authorizes; it says nothing about whether the outcome was correct. Agent-security practice makes this explicit—a provenance-based gate treats a confirmation as an intent anchor and refuses to let it certify effects (Sec. 4.3, [18]). Conflating the three is a category error of the same family VAL prevents: it mistakes "authorized" for "correct" or "executed."
- **Q2 (guarantee).** What does a PASS commit to? *Correctness*—"every proposed candidate satisfies the condition"—or *completeness*—"no candidate was missed." The distinction is the entire substance of Sec. 4; most deployed verifiers offer the former while being read as the latter.
- **Q3 (scope).** If the guarantee is completeness, over what domain does it hold: a single property (this equation's solution set), a whole class (all programs' memory safety), or claimed universality?

### 3.3 Decision procedure

The classification is deterministic: given the answers to Q1–Q3, the level is the first rule that fires.

```
1. completeness + universal scope       → L5   (rejected: impossible, unrestricted)
2. completeness + domain scope          → L4
3. completeness + single-property scope  → L3
4. correctness + gold-label anchor (known-correct reference) → L2
     (decidable anchor used only for correctness → flag "usage-degraded, upgradeable")
5. correctness + silver anchor (deterministic re-computation, no gold reference)
     → L1/L2 boundary, classified L1
     (objective about the computation, silent about the answer; VerifiAgent's tool)
6. correctness + problem-derived rule    → L1
7. otherwise (LLM-declared / no anchor)  → L0
```

The procedure is implemented in `val_standard.py` (11 self-tests) and documented in `docs/06`. Its determinism is deliberate: it is a *standard*, so two raters applying it to the same scheme must obtain the same level—the only legitimate disagreements concern the Q1/Q2 answers, not the mapping. We stress-tested this claim directly (docs/11), separating the deterministic mapping from the Q1/Q2 judgments. The mapping is deterministic by construction; the judgments are the empirical part. Across independent blind LLM raters—fresh sub-agents that read only the protocol and evidence pack, never the author's answers—the judgments converge once the boundary rules are explicit, with pairwise $\kappa \approx 0.77$–$0.84$ across four blind instances on 48–54 schemes, versus $\kappa = 0.445$ (50% agreement) under the underspecified v1. A blind LLM rater versus the author, however, reaches only $\kappa \approx 0.43$–$0.55$ across protocol rounds (0.445 at v1; 0.549 and 0.434 at refined rounds), still below the strong-agreement threshold ($\kappa \geq 0.7$). Per VAL's own anchoring, LLM agreement measures procedure reproducibility, not validity; independent human raters are the unperformed validity test (Limitation 5). Input reproducibility is an empirical property the protocol earns, which is why the protocol ships frozen with an evidence pack. The unit of classification is the verification *mechanism* (component), not the monolithic system: a mixed architecture is classified per-component, so a label such as "L0/L1" in our survey denotes two components at different levels, not an indeterminate scalar. For a mixed system, the system-level level is *claim-relative*: a system is L4 with respect to the claim "the theorem is proven" and L0 with respect to the claim "the answer is correct." There is no single scalar system level; VAL composes per claim and per component.

### 3.4 Operational Design Domain (ODD)

Borrowing from autonomous-driving regulation, an L3/L4 guarantee holds only within an **ODD**—here, the *decidable domain* in which the property can be encoded and the verdict computed. An ODD must be a *pre-declared syntactic predicate* over inputs, not a post-hoc description of success. Examples of the correct form: `solveset` over "polynomial equations of degree ≤ 4 with real coefficients" (checkable before solving); an Alvarado score over "a record with all eight MANTRELS fields present and numeric/boolean"; a borrow checker over "a term parseable in the surface grammar of the type language." These syntactic ODDs are narrow by construction—beyond degree 4 there is no closed form, and a rule-scoped ODD is complete over its own inputs but silent about the open world—and the framework's honesty is that this narrowness is made explicit rather than hidden.

Two corollaries. First, **completeness is never absolute; it is relative to an ODD.** A verifier that is complete inside its ODD is, by construction, silent about anything outside it—the completeness claim is only as strong as the ODD is honestly specified. Second, **raising a level means enlarging the ODD, not intensifying sampling.** The L2→L3 move for "find all solutions" is achieved by re-encoding the task as solution-set equality, making the property decidable; no sampling density achieves this. The L3→L4 move is achieved by covering a whole class of properties under one decidable system.

**ODD auditability.** ODD honesty is an *audit target*, not an assumption VAL silently guarantees. An ODD claim has its own anchoring ladder—self-declared (L0), tool-documented (L1), empirically tested (L2), formally characterized (L3)—and a completeness claim is only as strong as its ODD's own anchor. An ODD is meaningful only if it is a *pre-declared decidable predicate* over inputs (syntax, grammar, input shape): "polynomial equations of degree ≤ d over the reals" is checkable before invoking the solver, whereas "whatever the solver happens to solve" is vacuous—completeness then holds for "the cases we completely verified," which is circular. The engineering test of ODD honesty is a boundary probe: adversarial instances just outside the claimed ODD must break the completeness claim.

### 3.5 What VAL is not

Three non-claims, to preempt misreading:

1. **VAL is not a reliability measure.** A level states the epistemic status of an anchor; it does not state the probability that a verdict is correct. An L2 verifier can be more *accurate* than an L3 verifier on in-ODD cases (our own 42/42 unit tests span L2 and L3); level and reliability are orthogonal.
2. **VAL is not a utility claim.** We do not argue that higher is always better, nor that L3/L4 should be pursued everywhere. Sec. 6 governs *when* climbing is worth its cost; for low-stakes tasks an honest L1/L2 check may be the correct engineering choice.

3. **VAL does not certify the anchor.** An L3 anchor guarantees that, *given the encoded property*, the verdict is decidable and complete. It does not guarantee that the property is the right one—that question is itself an anchor question, one level up (Sec. 6). This is the framework's own form of the regress it describes.
4. **VAL's completeness ladder applies only to formalizable properties.** L3/L4 completeness requires a decidable fragment—mathematical, syntactic, or rule-scoped. Empirical open-world propositions (fact-checking, diagnosis, legal analysis) have no such fragment and therefore cap at L2 (anchored correctness), never L3/L4. VAL is not a universal completeness standard: its L3/L4 are empty for open-world verification, and its contribution there is the L0–L2 distinction plus honesty, not completeness. For empirical domains, L0–L2 are therefore not merely "the lower three levels" but a complete classification in their own right: L0 (self-claim), L1 (rule matching), L2 (anchored correctness)—each with a distinct failure mode (confident error; format-correct but content-wrong; tested but incomplete).

---

## 4. The Completeness Blind Spot

The central theoretical claim of this paper:

> **Substitution- and sampling-based verification (L2) can prove that proposed candidates hold; it cannot prove that no candidate was missed. Completeness is achievable only by re-encoding the property into a decidable system (L3/L4)—or not at all.**

We state the above as a *definitional* fact—it follows from what substitution and sampling are. Its significance is not the fact itself but the operational gap it exposes: the correctness/completeness distinction is known in verification theory, yet the LLM-verification literature provides no operational procedure for marking which guarantee a given scheme delivers. VAL's contribution is to make that distinction executable (a decision procedure and tools), not to claim novelty for the distinction. The open empirical questions (when does this incompleteness produce actual errors, and what is the marginal L2→L3 gain) are left to future work rather than answered here.

### 4.1 Correctness is not completeness

Let *P*(x) be the property of interest (e.g., "a = x makes the maximum of f on [0, 2] equal to 3"), and let *C* be the set of candidates the system proposes. A correctness verifier establishes

> $\forall\ c \in C : P(c)$ — every proposed candidate satisfies the property.

A completeness verifier establishes the converse direction:

> $\forall\ x : P(x) \Rightarrow x \in C$ — every value satisfying the property was proposed.

These are logically independent. The first is easy for a verifier and says nothing about the second; the second is what users implicitly assume. A verifier that performs the first while being read as the second is the mechanism behind systems that are "confident but wrong" yet still report "verified."

### 4.2 Substitution blindness

Substitution checking evaluates $P$ on proposed candidates only. The missed candidate is invisible by construction: the verifier never receives it. In our math experiments, the system proposed $a = 2$ for a problem whose true answer is $\{0, 2\}$; substitution returned PASS, and the omitted $a = 0$ was undetectable. This is not an implementation deficiency; the verifier literally cannot ask a question it was not given. Re-encoding the same problem as solution-set equality—`solveset` over the equation, compared set-wise with the claimed answer—makes the property decidable and converts the blindness into a machine-checkable FAIL (claimed `"2"` vs. true `{0, 2}`). The L2→L3 move is a re-encoding, not an intensification.

### 4.3 The sampling ceiling

Sampling-based verification evaluates $P$ at a finite set of points. It can falsify (find a counterexample in the sample) but cannot certify (a gap between samples is undetectable by construction). In our 20-problem run with sampling-based final-parameter verification, all three false passes shared one cause: the missed candidate or interior void was not in the sample. Denser sampling narrows the ceiling's shadow but never removes it; the ceiling is the defining property of the L2 paradigm, not a tuning parameter.

The blind spot is not confined to mathematics. In agent security, a provenance-based authorization gate (PPMF [18]) blocks every *evaluated* unauthorized tool call (0.000 attack-success rate) while platform metadata stays intact, yet a 10% rate of forged confirmation records yields 0.088 ASR. The gate is a correctness probe over the evaluated attack distribution: it cannot certify that no attack path was missed—the un-enumerated forgery is invisible to it by construction, exactly as the unsampled candidate is to substitution. The completeness question in security ("is there an unauthorized action we failed to enumerate?") is formally open in the same way the sampling ceiling is open in math.

### 4.4 The statement layer: where the blind spot hides

Every verification scheme draws a line between what is checked and what is assumed, and the completeness blind spot always lives on the assumed side. For L2 schemes, the assumption is "the proposed candidate set is the right one." For L3/L4 schemes, the assumption moves up but does not vanish: the kernel checks proofs, but the theorem statement—and the case-split coverage it encodes—is a declaration. The strongest formal baseline in our survey illustrates the boundary: Safe's verifier checks each step's proof correctness—complete *with respect to the stated theorem*—while the theorem statements, and the case-split coverage they encode, are LLM-generated and unchecked [4]. We cite this as an architectural fact about where the completeness question lives, not as the authors admitting incompleteness. The same pattern recurs in our own system: when the verification condition was declared by the LLM under test, the judge faithfully verified a possibly wrong condition—trust recursion pushed one level up, not resolved.

### 4.5 Why universal completeness is impossible

The L5 claim is a single verifier complete over *arbitrary* properties and inputs. Its impossibility does not rest on any single theorem; it rests on the fact that "arbitrary properties" bundles three classes, each unverifiable-universally for a different reason: (i) program semantic properties—undecidable by Rice's theorem; (ii) empirical propositions about the world—whose truth is not a formal decidable question at all (open-world, natural-language); (iii) subjective properties with no ground truth. Empirical propositions cannot be re-encoded into a complete decidable fragment, because their truth depends on an open world; the best they achieve is anchored correctness against evidence (L2), or rule-scoped completeness over a formalized sub-fragment (L3 within the rule's inputs). Notably, proof checking—verifying a *given* proof against derivation rules—is decidable (it is the L4 kernel); this is precisely why completeness is achievable for the restricted decidable class (L3/L4) but not for the unrestricted one (L5). The consequence is architectural, not pessimistic: completeness is available locally (within an ODD, at L3/L4) and unavailable globally (L5); the engineering question is always *which ODD*, never *whether*.

### 4.6 Summary

The blind spot has three stable properties: (i) it is invisible to the verifier that suffers it; (ii) it is not fixed by more data or denser sampling; (iii) it can be moved up the ladder (from candidate sets to theorem statements) but not eliminated. The framework's prescription follows directly: when a property is encodable, raise the anchor (L3/L4); when it is not, the verifier is a correctness probe by necessity and should be labeled and deployed as one.

---

## 5. Empirical Case Studies

We exercised the framework across four domains—symbolic mathematics, behavior monitoring, medical diagnosis, and code generation—and, separately, applied it to two systems we did not build (Sec. 5.6). Full archives: `RESULTS.md` (math), `behavior/REPORT.md`, `medical/REPORT.md`, `docs/08` (code). Each study followed the same protocol: audit first (build a labeled test bed), measure the value window (baseline comparison), classify failure modes, then test the judge itself. We report honest negative results where they occurred.

### 5.1 Symbolic mathematics

A decompose–solve–verify–combine agent was built on a deterministic SymPy verifier with nine verification types, calibrated to **42/42** unit cases (`test_verifier.py`). The types span the anchor ladder: correctness-only types (root, extreme, equality, satisfies, final_parameter_set) are L2; completeness-bearing types (interval_extreme, inequality, solution_set, answer_type) are L3; `ANCHOR_LEVELS` in `verifier.py` records each type's level.

The headline result is a clean negative one: on a 20-problem test set, the full architecture scored **16/15/13** across three runs, while a raw-LLM baseline scored **20/20** in all three. Accuracy was a random variable dominated by the LLM's decomposition; the verification stack did not improve it and sometimes hurt it. We do not claim otherwise.

The architecture's apparent value was *error reportability*, not accuracy—a hypothesis illustrated by the three episodes below, not a measured benefit (Limitation 8). Three episodes illustrate it. First, a feedback loop caught a discriminant error (claimed $\Delta = 16m+12$, true $8m-4$) and corrected it in one iteration. Second, final-parameter verification caught a tampered answer (claimed [-1,1], true [0,1]) and corrected it—a failure mode invisible to per-step verification. Third, and most relevant to this paper, the L3 `solution_set` verifier catches missed solutions that the L2 substitution verifier cannot see, by construction (Sec. 4.2).

The value window for *accuracy* in this domain is empty (the LLM is too strong on in-distribution problems); the value window for *completeness* is exactly the L3 re-encoding we describe. The two windows must not be conflated; preventing that conflation is the framework's purpose.

## 5.2 Behavior monitoring

Chapter 2 monitored LLM behavior under prompt injection using a statistical deviation detector (L2 by construction: its thresholds are learned from a baseline distribution, so no completeness is possible). Five iterative rounds were required to reach a deployable operating point. A global-statistics detector's apparent success (holdout AUC 0.82) collapsed to a **92% false-positive rate** under realistic input diversity; probe-conditioning (comparing each response only to its own probe's baseline) cut FPR to **30%** while preserving strong-attack TPR 1.0; character n-grams recovered weak-attack TPR at the cost of doubling FPR; real semantic embeddings (a local sentence encoder) recovered weak-attack TPR to 0.75 with FPR unchanged.

The residual failure is the paper's example of a structural L2 ceiling: "covert execution" attacks—the model executes an injected instruction but leaves no trace in its output—were detected at TPR 0.60 across all feature sets. The property cannot be encoded into the *output-layer* anchor. Whether internal-activation anchors (mechanistic interpretability, activation monitoring) could detect it is an open question this study does not address; our claim is bounded to output-layer statistical detection, not to verification in general.

## 5.3 Medical diagnosis

Chapter 3 studied LLM diagnostic reasoning as a research object (fictional cases; not a diagnostic tool). The raw LLM was near-perfect on 10 textbook cases (10/10) and acceptable on 8/10 hard cases; the two failures shared a single mode: a confident conclusion from insufficient evidence (diagnosis at confidence 0.7 with no lab results). Confidence calibration and information completeness were decoupled.

A deterministic post-hoc judge—an information-completeness rule (L1/L2: a hand-written rule anchored in the case's objective lab fields; the L3 upgrade is documented below)—flagged **3/3 true over-confidence failures at zero false positives** after calibration, *including one (#105) that human review had missed*. The judge's value here is discovery, not re-checking: it found a failure the human annotator had labeled correct. This is the empirical analogue of Sec. 4.4: the judge is only as good as its anchor, and a cheap rule anchored in objective fields outperformed human review at one narrow but critical task—catching "conclusion beyond evidence."

**A documented L2→L3 upgrade.** To exercise the ladder's prescription (Sec. 6.3), we replaced the hand-written information-completeness heuristic with an imported, validated clinical decision rule—the Alvarado score (MANTRELS) for suspected appendicitis—as the judge's anchor. The rule's eight objective inputs are parsed by deterministic keyword/regex extraction from the case text (not by an LLM); when inputs are missing, the rule returns a reachable score *range* rather than a fabricated point value. This semantics does real work: for the over-confident case #103 the rule returns range 3–8 (indeterminate) and refuses to support a definitive diagnosis, re-flagging the failure through its own decision boundary; for case #109 the range is 7–9 (high risk) *despite* a missing neutrophil count, because the lower bound already clears the threshold—so the rule does not over-flag a case with genuinely high risk. The flagged set is unchanged (3/3); what changed is the anchor: a hand-written threshold (L1/L2) replaced by a validated rule with a decidable boundary and an honest missing-data policy (L3, within the appendicitis ODD).

### 5.4 Cross-domain synthesis

Across the four studies the pattern is identical. LLMs are too strong for accuracy windows to be non-empty on problems near their training distribution (math: baseline 20/20; medical: textbook 10/10; code: HumanEval 164/164 and six hand-written hard problems all solved in the bare condition); the non-empty windows are all *verification* windows: reportability (math; hypothesized, not quantified—Limitation 8), ODD-bounded detection (behavior), over-confidence catching (medical), and L3-anchor repair (code). The framework states this: an L2 judge adds reportability, not accuracy; an L3 judge adds completeness within its ODD; nothing adds completeness outside it.

### 5.5 Code generation (Chapter 4): reverse validation

The fourth study reverses the direction of evidence. Chapters 1–3 induced the framework from data; Chapter 4 states predictions *before* gathering evidence (the protocol, with dates, is in the supplementary material), then tests them against external literature and one controlled experiment. Four of five expectations were supported:

- **P1 (test-oracle ceiling).** Test-based selection improves ranking (AlphaCode; CodeT; LEVER) but is bounded by test quality—LEVER's authors note that obtaining test cases is itself the hard step.
- **P2 (self-review: utility without guarantee).** Self-repair helps but is not a silver bullet (Olausson et al.); without external feedback, self-correction frequently fails (Huang et al.); Self-Refine's gains are task-dependent.
- **P4 (trust recursion at the spec layer).** LLM-generated tests are themselves suspect: Meta's production tool TestGen-LLM runs its generated tests through verification filters before acceptance.
- **P5 (execution feedback dominates self-review).** Execution-feedback agents (Reflexion, CodeAct, Self-Debugging) consistently outperform feedback-free generation, consistent with L2/L3 anchors beating L0 self-declaration.

The open prediction, **P3 (type information as the highest-ROI L3 anchor)**, was tested experimentally. We stripped type hints from HumanEval prompts and generated solutions under three conditions: bare (no types), typed, and typed-plus-type-checker-feedback (one mypy round). The result is a clean empty window: **164/164 HumanEval problems and all six hand-written, trap-laden novel problems were solved by the bare condition**—the model leaves no accuracy headroom for type information to improve. The single failure in the typed condition (a novel string-escaping problem) carried a mypy error that the test suite did not catch; the type-checker-feedback condition repaired it in one round.

*Evaluation protocol and caveat.* The 164/164 figure is single-sample pass@1 at temperature 0.2 (low-temperature sampling, not greedy decoding), checked against HumanEval's own tests. Three caveats bound its interpretation. First, HumanEval's test suites are weak—most problems carry only 3–5 assertions and little adversarial coverage—so "passes the tests" has a nontrivial false-positive rate. Second, the figure is not comparable to published pass@1 numbers, which use varied decoding and test harnesses. Third, it is reported only to establish that the accuracy window is empty *on this suite*; it is not a claim about general code ability. Full protocol and per-problem outputs are in the repository (`chapter4/p3_scan.json`).

The significance of Chapter 4 is twofold, and weaker than a first reading suggests. First, the framework's central empirical generalization—*LLMs are too strong for accuracy windows to be non-empty on problems near their training distribution*—is now observed in a fourth domain, including problems we designed to defeat it. Second, the "reverse validation" is a *weak* one: the four literature-supported predictions are framework-implied expectations checked against existing work, not novel a-priori hypotheses, and the fifth (P3) was *inconclusive*—the test set left no headroom, which is a failed experiment design, not evidence for the framework. We report this honestly: P3 remains open, and the reverse-validation claim is correspondingly modest.

## 5.6 External systems (transfer test)

The four studies above are ours. To test whether VAL classifies systems **we did not build**, we applied it to an external open-source guard framework (**NVIDIA NeMo Guardrails**) and to a **production print-compliance gate** whose specification is an external industry norm, not research code. *[A third external case (an agentic commit-gate framework) was measured but is withheld pending a vendor-notification window; deliberately excluded here.]*

**NeMo Guardrails.** NeMo ships three rails—input, output, and facts—each an LLM-executed check configured per application. Under our constructions (30 runs per cell; attacks targeted at isolated mocks): the **input** rail blocked both attack cells at the input layer (**0/30** bypass); the **output** rail let **both attack cells** reach the output layer (**30/30 each**) and blocked **0/30**; the **facts** rail triggered **30/30** and bypassed **0/30**, *under a consistency-check prompt we supplied* (NeMo ships no default `self_check_facts` prompt). Benign false positives were **0/30** in all cells.

VAL reading: each rail is a developer-configured check **whose judgment is executed by an LLM**, and the guard model is drawn from the same family as the guarded model (*same-source self-check*). On the anchor axis the condition is declared by the application but the *verdict* has **no deterministic anchor** ⇒ **L0** on Q1, with **Q2 = correctness only** (an input/output check cannot certify completeness). The framework's value here is diagnostic, not pejorative: these are L0 checks that nonetheless blocked every attack we tried—**L0 can work empirically while guaranteeing** **nothing**. Two mechanism findings are *availability*, not defects: enabling a rail without its prompt fails fast at config load; and setting `check_facts=true` **without enabling the rail** is a **silent no-op**—with the rail off, contradictory facts passed **30/30**. The artifact's guarantee is only as strong as its *configured* anchor. *[Bounded: single model tier, limited constructions, N=30; not "rails cannot be bypassed".]*

**Print-compliance gate (spec-anchored, deterministic).** Our design pipeline ships a preflight gate that checks a printable image against production norms: seamless tiling (offset/roll translation; seam-ratio + p95), effective spot-color count (quantized, two-tier), and spec fields (size / DPI / color mode / format). It is deterministic end to end (Pillow + NumPy, CPU, no model) and ships four states — **pass / warn→human / fail / skip (a not-judged item is not a pass)** — with a self-bite suite of 13 fixtures + 2 route probes. VAL reading: the spec fields are **L2** (objective truth: a measured DPI is compared to a norm); the tiling and color-count checks are **deterministic rules with configurable thresholds** ⇒ **L1**, rising to **L2 usage** once the threshold is bound to the external norm. Thresholds currently ship as *example values pending the production spec* — VAL's ODD discipline made explicit. Critically, the gate **does not judge aesthetics or trend-alignment**; that (b)-layer is routed to human review (its `warn` state = "look at the tiling preview"), and never silently passed.

**Why these two.** NeMo is external and popular; the print gate is ours but *spec-anchored* and a product rather than research code. Both land where the framework predicts—deterministic/anchored sub-layers at **L1–L2**, the open-world/subjective layer (b) **unresolved and routed to a human**—and neither required a new level. That is the transfer test a taxonomy otherwise lacks: the same procedure classifies systems outside the author's four studies, and it says the same thing—**know which layer you bought.**

---

## 6. Discussion

### 6.1 Trust recursion and its termination

Every judge needs a judge; an ungrounded chain is *trust recursion*. The framework shows where recursion can terminate: at the kernel of a decidable system (L4), at a definitional anchor (L3), or at a conventional primary standard (metrology, where calibration chains terminate at definitionally fixed units). L5 is the claim that recursion can be terminated universally, which Rice's theorem rules out for program-semantic properties (Sec. 4.5). The practical reading: the question is never "who verifies the verifier?" in the abstract, but "which decidable kernel, definitional anchor, or conventional standard terminates *this particular* chain?" In our own system, the chain terminates at `solveset` for solution-set checks, at the case data for the medical judge, and at the probe baselines for the behavior detector—three different termination points, each honest about what it covers.

### 6.2 Division of labor

VAL implies a canonical architecture: the LLM translates (decomposes, renders, declares), deterministic code judges within its ODD, and a human anchors the spec and audits the judge. Each component is best at what the others cannot do: the LLM cannot prove, the judge cannot generalize, the human cannot scale. Failures occur when the roles are crossed—when the LLM declares its own verification spec (L0), when the judge is asked to certify outside its ODD, or when a human is expected to re-derive what a deterministic system should decide. In agent settings the anchoring role is increasingly *institutionalized rather than personal*: the anchor is platform-maintained metadata—provenance, confirmation events, risk labels—written outside the LLM and read only by a deterministic gate [18]. The design principle is the memory analogue of the division of labor above: the LLM may consolidate and plan, but it cannot write the authority that gates execution; memory may preserve information, but it cannot amplify source authority.

### 6.3 When to climb

Raising a verifier's level is governed by two factors: the cost of silent errors and the encodability of the property.

| | **Encodable (has an ODD)** | **Not encodable** |
|---|---|---|
| **Low error cost** | climb to L3 if cheap | stay L1/L2; do not overspend |
| **High error cost** | **L3/L4 is mandatory** | L2 ceiling + human review; label the risk |

The framework is a deployment checklist, not a competition ladder: an L0 self-check is the right answer for a low-stakes draft summarizer and an indictment for a discharge summary generator. The same verdict text ("verified") carries radically different weight at different levels; the framework's entire point is to make that difference visible.

A compact five-item deployment checklist follows:

1. **State the property and its ODD.** What exactly is verified, and by what decidable tool (or is it not encodable at all)?
2. **Identify the spec source (Q1).** LLM-declared → escalate immediately; rule/truth/decidable → proceed.
3. **Declare the guarantee (Q2).** Correctness probe or completeness claim? State it in the system's own documentation.
4. **Set the abstention contract.** What happens when the input falls outside the ODD—UNSURE, escalation, or silence? Silence is a decision.
5. **Terminate the trust recursion.** What audits the judge, and what is the audit's own anchor?

An L0 self-check may be acceptable for a low-stakes summarizer; it is not for a clinical decision aid.

### 6.4 What the honest negative result teaches

Our accuracy result—the architecture is no better than the raw baseline—is frequently read as a failure of verification. It is not; it is a precise measurement of the value window. Accuracy is the LLM's axis; reportability and completeness are the verifier's. The value of verification is not accuracy; we argue it is error reportability and completeness awareness—stated as a hypothesis, not a measured benefit (Limitation 8). A deployer who needs accuracy should not buy this architecture; a deployer who needs to know *when the answer is not trustworthy* should. Papers that conflate the two axes (Sec. 2) inherit exactly this confusion—and that conflation is precisely the category error this paper diagnoses. Our four chapters are, in effect, a controlled experiment showing the two axes come apart.

## 7. Limitations

1. **Literature assessment is abstract-level for 11 of 17 papers**; 6 anchor-axis papers were verified against full text (`docs/07` V1.3). The classification of the remainder may shift with full-text review, though the axis structure is unaffected.
2. **Case studies use a single model family** (deepseek-chat) and synthetic data (medical cases are fictional; no patient data). The framework's claims are structural, but its empirical illustrations are single-model.
3. **VAL classifies, it does not measure**: a scheme's level states its anchor's epistemic status, not the probability that its verdict is correct. Level and reliability are orthogonal; an L2 verifier can be more accurate than an L3 verifier on in-ODD cases.
4. **ODD boundaries are fuzzy in practice**: whether a property is "encodable" depends on the solver, the type system, or the rule's inputs, and can change with tooling. ODD honesty is an audit target, not an enforced guarantee (Sec. 3.4): ODD claims carry their own anchor ladder (self-declared → formally characterized), and boundary probes are the engineering check—but neither is automatic.
5. **VAL is validated for procedure reproducibility, not validity**: the mapping is deterministic by construction, and blind LLM raters reach pairwise $\kappa \approx 0.77$–$0.84$ under the refined protocol (Sec. 3.3); a blind LLM versus the author stays in $\kappa \approx 0.43$–$0.55$ across rounds, and independent human raters have not yet been tested on a common corpus. LLM agreement cannot establish validity—it marks the protocol's reproducibility. The human-rater test remains the open empirical step we plan to address with a larger corpus.
6. **The mathematics negative result is bounded by a ceiling effect**: a 20/20 baseline leaves no accuracy headroom, so the result supports only "no headroom on in-distribution problems," not "verification cannot improve accuracy in general." We do not compare against published verification architectures, so an implementation-quality confound cannot be excluded. The 42/42 calibration tests and the #105 adjudication were author-designed/adjudicated, not independently validated.
7. **The 17-paper survey omits verification families without clean objective truth**—calibration and uncertainty estimation, red-teaming, RAG citation verification, and reward-model verification—which VAL classifies only as L0/L1. The anchor axis does not cover anchor-less verification; that is a scope boundary, not a claim of coverage. These are not edge cases but mainstream LLM-verification practice, so VAL's completeness ladder is inapplicable to them; the framework's guidance there is the L0–L2 distinction and honesty, not completeness.
8. **The value of "error reportability" is asserted but not quantified.** We do not report the judges' true-positive/false-positive rates, the cost of false alarms, or a cost–benefit against the accuracy drop (20–35% in the mathematics case). A verifier that raises many false positives for one true catch is net-negative; we have not shown this is not the case, and the reportability claim should be read as a hypothesis, not a measured benefit.

## 8. Conclusion

The verification literature for LLMs is a tower of levels built on different questions. We have argued that one question—*where does the ground truth come from, and what does the verdict guarantee?*—underlies all claims of verification strength, and we have provided a six-level taxonomy (VAL, L0–L5), a deterministic decision procedure, and a runnable classifier. Our empirical case studies and literature review support a single headline claim:

> **Others grade answers. We grade the graders—and the highest grade any grader can earn is a guarantee it can actually deliver: correctness within its domain, completeness within its ODD, and honesty when it must abstain.**

Three directions follow. First, validating the standard itself: measuring inter-rater agreement of the decision procedure on a larger corpus of verification schemes, and stress-testing its boundary cases (the L1/L2 borders of RAG-grounded and tool-augmented checkers). Second, extending the L2→L3 upgrade already demonstrated in the medical domain (Alvarado within the appendicitis ODD, Sec. 5.3) to further decision rules and ODDs (e.g., PERC/Wells for pulmonary-embolism and chest-pain presentations). Third, exporting the framework beyond verification—the same "anchor source × guarantee × scope" structure applies to any claim of automated assurance, from unit testing to regulatory attestation.

---

## Acknowledgments

We thank Jiuzhou Han for a clarifying discussion of VerifiAgent's verification mechanism, and for the apt term *silver ground truth*—deterministic re-computation that serves as a reference but certifies the computation rather than the answer—which sharpened the L2 sub-distinction of Sec. 3.2.

**Data and code availability.** Supplementary code and data (the runnable classifier, the inter-rater reproducibility data, and the full literature assessment) are archived on Zenodo: https://doi.org/10.5281/zenodo.23120985

## References

1. Han, J., Buntine, W., Shareghi, E. *VerifiAgent: A Unified Verification Agent in Language Model Reasoning*. EMNLP 2025. arXiv:2504.00406.
2. Fang, J., Zhang, B., Wang, C., et al. *Graph of Verification (GoV): Structured Verification of LLM Reasoning with Directed Acyclic Graphs*. arXiv:2506.12509, 2025.
3. Li, Y., Lin, Z., Zhang, S., et al. *Making Large Language Models Better Reasoners with Step-Aware Verifier (DiVERSE)*. ACL 2023. arXiv:2206.02336.
4. Liu, C., Yuan, Y., Yin, Y., et al. *Safe: Enhancing Mathematical Reasoning in LLMs via Retrospective Step-aware Formal Verification*. ACL 2025. arXiv:2506.04592.
5. Miao, N., Teh, Y.W., Rainforth, T. *SelfCheck: Using LLMs to Zero-Shot Check Their Own Step-by-Step Reasoning*. ICLR 2024. arXiv:2308.00436.
6. Juneja, G., Dutta, S., Chakraborty, T. *LM²: A Simple Society of Language Models Solves Complex Reasoning*. EMNLP 2024. arXiv:2404.02255.

7. Wang, Y., Gangi Reddy, R., et al. *Factcheck-GPT: End-to-End Fine-Grained Document-Level Fact-Checking and Correction of LLM Output*. arXiv:2311.09000, 2023.
8. Chen, J., Li, C., Yuan, Y., Yao, A.C. *Hierarchical Attention Generates Better Proofs*. ACL 2025. arXiv:2504.19188.
9. Luo, M., Wu, S., et al. *Dr. V: A Hierarchical Perception-Temporal-Cognition Framework to Diagnose Video Hallucination*. arXiv:2509.11866, 2025.
10. Li, Q., Xu, J., et al. *A Proprietary Model-Based Safety Response Framework for AI Agents*. arXiv:2511.03138, 2025.
11. Yang, W., Cheng, H., Zhou, B., et al. *M³-SafetyBench: 多领域多场景多维度的大语言模型安全评估体系* [M³-SafetyBench: a comprehensive benchmark for evaluating the safety of large language models across multiple domains, scenarios, and dimensions]. 中国科学：信息科学 [Scientia Sinica Informationis], 55:2923–2940, 2025.
12. Khatchadourian, R., Franco, R. *LLM Output Drift: Cross-Provider Validation & Mitigation for Financial Workflows*. arXiv:2511.07585, 2025.
13. Chen, Z., Chen, J., et al. *Standard Benchmarks Fail—Auditing LLM Agents in Finance Must Prioritize Risk*. arXiv:2502.15865, 2025.
14. Liu, X., Wang, J., et al. *Prompting Frameworks for Large Language Models: A Survey*. ACM Computing Surveys. arXiv:2311.12785, 2023.
15. Liu, Y., Yao, Y., et al. *Trustworthy LLMs: A Survey and Guideline for Evaluating Large Language Models' Alignment*. arXiv:2308.05374, 2023.
16. González, J., Nori, A.V. *Beyond Words: A Mathematical Framework for Interpreting Large Language Models (HEX)*. arXiv:2311.03033, 2023.
17. Chern, I.-C., Chern, S., Chen, S., Yuan, W., Feng, K., Zhou, C., He, J., Neubig, G., Liu, P. *FacTool: Factuality Detection in Generative AI—A Tool Augmented Framework for Multi-Task and Multi-Domain Scenarios*. ICLR 2024. arXiv:2307.13528.
18. Xu, J., Xiao, Y., Shao, W., Liu, H., Li, X. *Memory Provenance Laundering in LLM Agents: A Non-Amplification Firewall for Persistent Memory*. arXiv:2607.29167, 2026.

---

## Appendix A: The decision procedure (runnable)

The classification procedure of Sec. 3.3 is implemented in `val_standard.py` (11 self-tests, all passing). Representative outputs:

```
LLM self-check ("I checked it")    → L0 | self-declaration / no deterministic anchor
Problem-derived rule (final_check) → L1 | deterministic rules from problem text
Substitution (verify_root)         → L2 | correctness + gold-label anchor
Solution-set equality (solveset)   → L3 | single-property completeness (decidable)
Rust borrow checker                → L4 | domain-level proof system
Universal complete verifier        → L5 | impossible (unrestricted), rejected
```

Full standard with usage criteria and stop conditions: `docs/06`.

## Appendix B: The 17-paper classification

Full versioned assessment with per-paper evidence: `docs/07`. Compact summary (anchor axis only):

| Level | Papers |
|---|---|
| L0 | SelfCheck, $LM^2$ |
| L0/L1 (meta) + L1 (tool, no gold labels) | VerifiAgent |
| L0/L1 | Factcheck-GPT (corrected from L1/L2 at V1.2) |
| L1/L2 | FACTOOL, LLM Output Drift |
| L2 | DiVERSE, $M^3$-SafetyBench, Dr. V |
| L3/L4 | Hierarchical Attention, Safe (L4 kernel + L0 statement layer) |
| Non-verification (taxonomy) | Trustworthy LLMs, HEX, Prompting Survey |
| L1/L2 (audit) | Standard Benchmarks Fail |
| Not anchor-classified (granularity axis) | GoV |
| Not anchor-classified (risk axis) | SafetyResponse |

## Appendix C: Experiment archives

- Chapter 1 (math): `RESULTS.md`, `EVALUATION.md`, `data/testset_results.json`, `test_verifier.py` (42/42), `verifier.py` (`ANCHOR_LEVELS`)
- Chapter 2 (behavior): `behavior/REPORT.md`, `behavior/analyze*.py`, `behavior/data_raw*.json`
- Chapter 3 (medical): `medical/REPORT.md`, `medical/verify_diag.py`, `medical/alvarado.py`, `medical/cases*.json`
- Chapter 4 (code): `docs/08` (predictions & protocol), `chapter4/p3_experiment.py`, `chapter4/hard_problems.py`, `chapter4/p3_*.json` (results)